\documentclass[10pt,twocolumn,letterpaper]{article}

\usepackage{cvpr}
\usepackage{times}
\usepackage{epsfig}
\usepackage{graphicx}
\usepackage{amsmath}
\usepackage{amssymb}
\usepackage{float}

\usepackage[breaklinks=true,bookmarks=false]{hyperref}

\DeclareMathOperator*{\argmax}{arg\,max}

\cvprfinalcopy 
\onecolumn
\def\cvprPaperID{****} 

\begin{document}
\title{Protecting JPEG Images Against Adversarial Attacks}

\author{Aaditya Prakash, Nick Moran, Solomon Garber, Antonella DiLillo and James Storer\\
Brandeis University\\
{\tt\small aprakash,nemtiax,solomongarber,dilant,storer@brandeis.edu}
}

\maketitle
\begin{abstract}
As deep neural networks (DNNs) have been integrated into critical systems, several methods to attack these systems have been developed. These adversarial attacks make imperceptible modifications to an image that fool DNN classifiers. We present an adaptive JPEG encoder which defends against many of these attacks.
Experimentally, we show that our method produces images with high visual quality while greatly reducing the potency of state-of-the-art attacks.
Our algorithm requires only a modest increase in encoding time, produces a compressed image which can be decompressed by an off-the-shelf JPEG decoder, and classified by an unmodified classifier.
\end{abstract}
\section{Introduction}
Deep neural networks (DNNs) have shown tremendous success in image recognition tasks, even surpassing human capability~\cite{He2016DeepRL}. 
DNNs have become components of many critical systems, such as self-driving cars~\cite{bojarski2016end}, medical image segmentation, surveillance, and  malware classification~\cite{pascanu2015malware}. However, recent research has shown that DNNs are vulnerable to adversarial attacks, in which minute, carefully-chosen image perturbations can result in misclassification of the image by the neural network~\cite{Goodfellow2014ExplainingAH}.
In most cases, this change is imperceptible to humans (the resulting image is visually indistinguishable from the original).


Current adversarial attacks take advantage of the \emph{over-completeness} of the pixel representation of images -- that is, storing images as an array of values results in storing much more information than is required to recognize the content of an image. Traditional image compression techniques, like JPEG, rely on assumptions about the human perception of natural images in order to compress them. Since adversarial attacks are imperceptible to the human eye, they can be viewed as hiding themselves in the over-complete part of the pixel representation of images. Removing these adversarial perturbations is therefore a lossy image compression problem: by effectively encoding natural images, image compression techniques can quantize away imperceptible elements of an image, effectively blocking attacks that rely on adversarial perturbations.


While the research in defending against such adversarial perturbations is still in its infancy, it has been shown that JPEG naturally removes some of these imperceptible  perturbations, thereby restoring the original classification of the image~\cite{Das2017KeepingTB,Dziugaite2016ASO}.
More advanced attacks, however, are robust against JPEG compression.
In this work, we study the nature of the perturbations generated by adversarial methods, as well as their effect on classification. We present a solution which produces a standard JPEG-decodable image, while allowing significantly improved classification recovery as compared to off-the-shelf JPEG compression.


Several techniques have been proposed which attempt to reduce the feasibility of generating adversarial images~\cite{Papernot2016DistillationAA,Miyato2015DistributionalSW}. Most of these methods involve augmenting the DNN training process to create a more robust classifier.
However, these defenses often fail against more advanced attacks, and come at the cost of more computationally expensive training. 

\begin{figure}
     \centering
     \includegraphics[width=0.30\textwidth,angle=0]{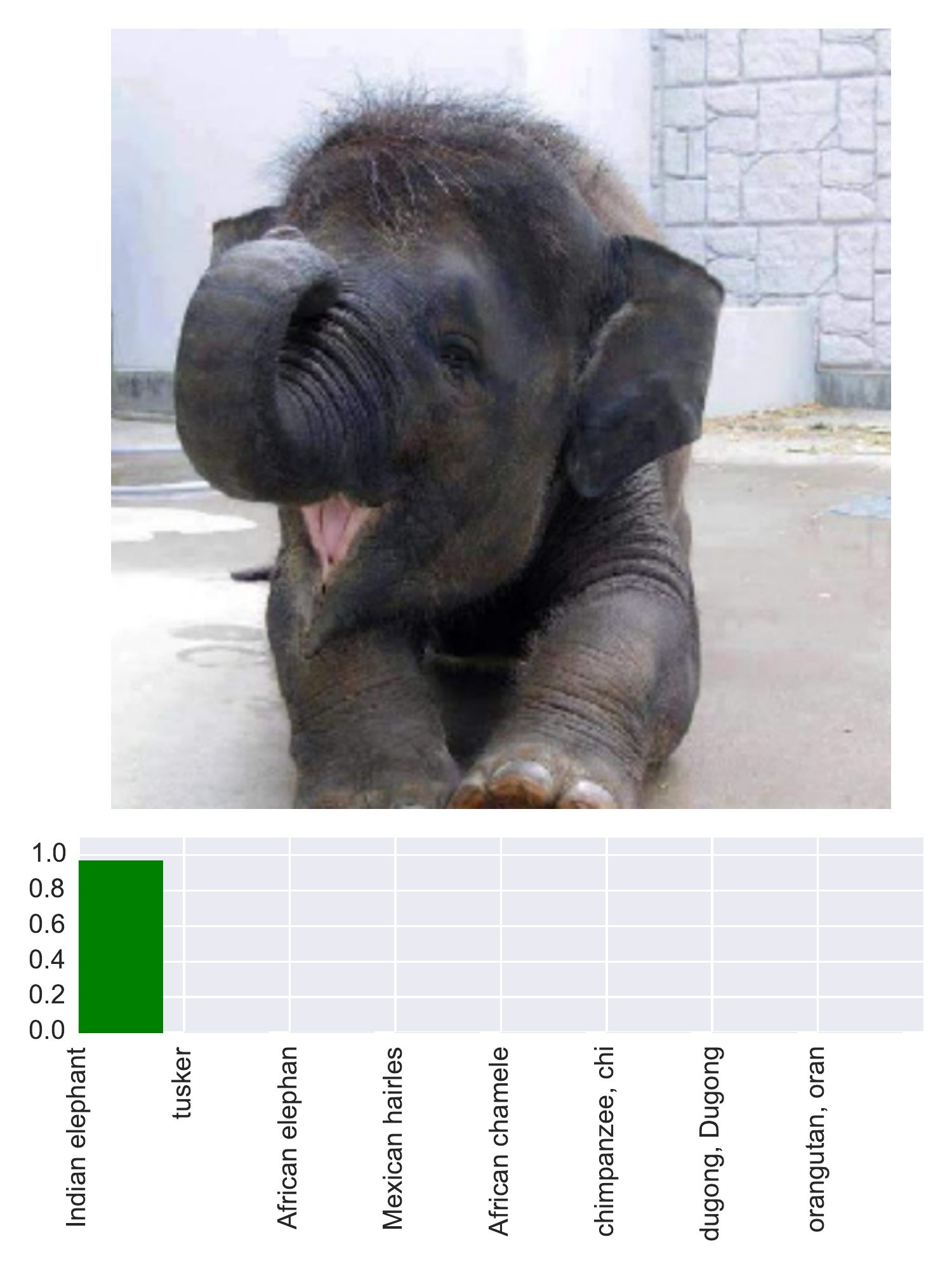}
     \includegraphics[width=0.04\textwidth,angle=0,scale=0.9]{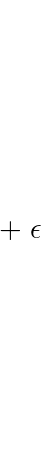}
     \includegraphics[width=0.30\textwidth,angle=0]{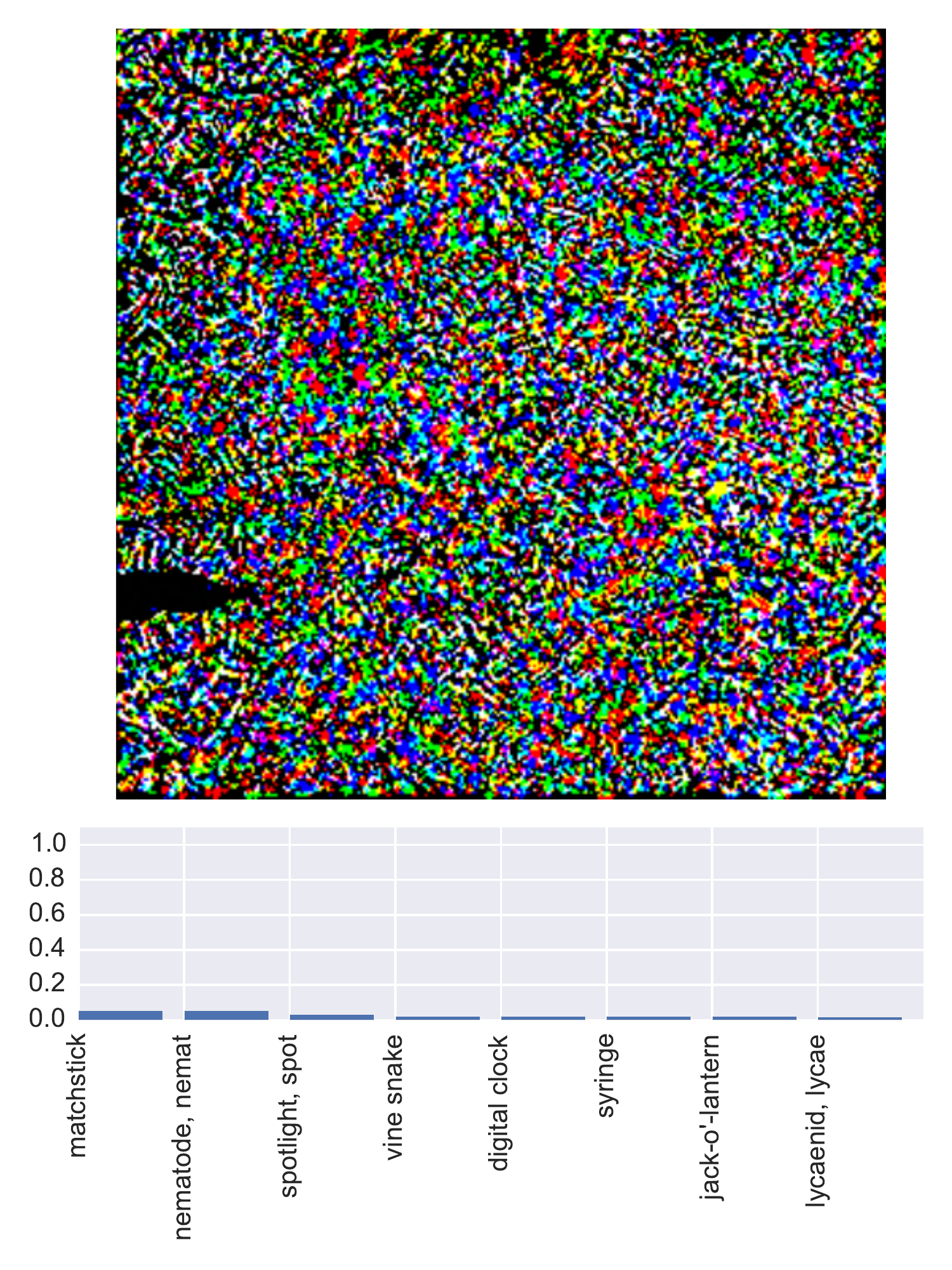}
     \includegraphics[width=0.03\textwidth,angle=0,scale=0.9]{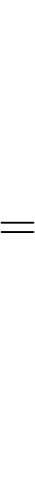}
     \includegraphics[width=0.30\textwidth,angle=0]{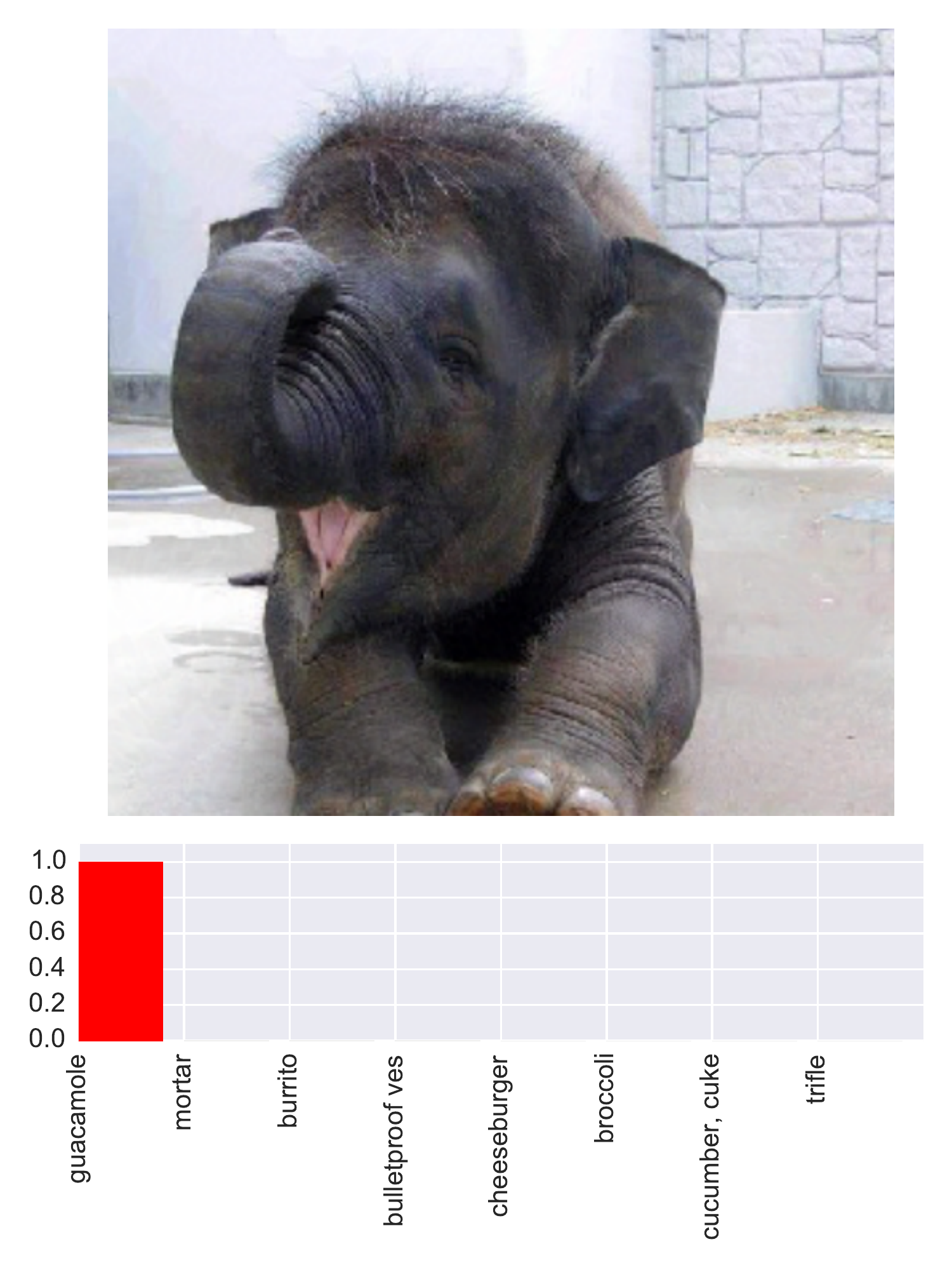}
     \label{fig:adv}
      \caption{Left: Original Image, Center: Perturbed Noise, Right: Adversarial Image}
 \end{figure}




Standard JPEG compression is able to provide some measure of defense against adversarial attacks.
Since adversarial perturbations are often high frequency and low magnitude, the quantization step of JPEG frequently removes them.
When the noise added by the adversary is removed or substantially modified, and the resulting image no longer fools the classifier.
However, as the severity of quantization increases, important features of the original image are also lost, and the model's ability to correctly classify the resulting image deteriorates regardless of any adversarial perturbation.
This trade-off limits the effectiveness of a simple JPEG-based defense, as defending against more sophisticated attacks requires harsh quantization that renders the output image difficult to classify.

Our algorithm employs an adaptive quantization scheme which intelligently quantizes different blocks at different levels. 
We have found that the regions chosen for perturations by attack methods do not strongly correlate with the presence of a semantic object.
As a result, there often exists added noise outside the object of interest in the image. We target this portion of the noise for aggressive quantization.
We tested the MSROI algorithm of ~\cite{Prakash2017SemanticPI}, which identifies multiple salient regions of the image, and quantizes those regions which are not salient more aggressively. 
MSROI adaptive quantization dampens a significant portion of the adversarial noise, but does so without deteriorating the quality of salient regions, thereby disrupting the attack while allowing for recovery of the original classification.
However, initial experiments demonstrated that the unmodified MSROI algorithm is also susceptible to adversarial noise, weakening the accuracy of the generated saliency maps.
Therefore, we present a more robust method of generating a saliency map, which is able to recover object saliency even in the presence of adversarial noise, without significantly impacting performance on unperturbed images.
We show that our proposed defense is successful against several state-of-the-art attack methods using an off-the-shelf classification model.
We further demonstrate that our proposed technique greatly improves upon the defense provided by a standard JPEG encoder.

\section{Background}
A deep neural network can be viewed as a parameterized function, $F_\theta$, which maps $x \in \mathbb{R}^n$ to $y \in \mathbb{R}^m$  i.e.\ $F_\theta(x) = y$ where $\theta$ represents the model parameters. 
Feed-forward neural networks are composed of multiple layers of successive operations, each applied to the output of the previous layer. For a deep neural network containing $l$ layers, the final network is a function composition of $l$ simpler functions i.e $F = F_1 \circ F_2 \circ \cdots F_l$.
These networks are trained by minimizing some loss function, $\mathcal{L}$, which is a differentiable measure of the classification accuracy on the training data.
The parameters of the model, $\theta$, are learned by stochastic gradient descent (SGD). 

Most adversarial attack methods use the gradient of the image pixels with respect to the the classification error to generate the desired perturbation.
In order to compute this gradient, an attacker needs to have access to the internal model and parameters.
These kinds of attacks are called white box attacks, as the adversary needs access to the targeted model. 
White box attacks can fool many state-of-the-art convolutional neural networks. 
Many popular deep learning models are publicly available, making them vulnerable in practice to these attacks.


It is also possible to attack deep networks without access to the model parameters. 
Such attacks are called black box attacks, and they begin by training a secondary model to match the outputs of the target model.
This secondary model can then be attacked using white box methods, and the resulting adversarial images are transferable to the target network \cite{Papernot2016PracticalBA}. 
Black box attacks are weaker than white box attacks; here we focus on white box attacks. 

\section{Adversarial Attacks} \label{attacks}
Formally, the paradigm of adversarial attacks can be described as follows. 
Let $x$ be an input image to a classifier. 
Then, the class label (the model's evaluation of what object the image contains) for the image $x$ is: $c = \argmax_{i} F(x|\theta)$ where $i$ is an index into the output vector.

An adversarial attacker takes the image $x$ and adds a small perturbation $\delta_x$, to obtain an adversarial image $x' = x + \delta_x$.
The attack is considered to be successful if, for a small $\delta_x$, the class label of the adversarial image ($c'$), is not the same as that of original image ($c$). 
There is no consensus on the exact definition of  ``small'' for $\delta_x$, but attacks are generally considered successful when the difference between $x$ and $x'$ is indistinguishable to the human eye. Some attacks are noticeable to humans, but do not change our perception of the class label of the image, and these attacks are also considered successful.  
Attacks can be targeted to induce a specific chosen class label, $\hat{c}$ such that $c' = \hat{c}$, or untargeted, meaning that they seek to induce any class label, as long as it does not match the original classification, i.e., $c' \neq c$.

In most cases, untargeted attacks are performed by iterating a targeted attack over all classes and choosing the class which requires the smallest perturbation.
As our work focuses on defense, we will consider untargeted attacks, and therefore our methods should also be effective against targeted attacks.

Adversarial attacks work by exploiting the design of the model and the limitations of SGD. 
Just as gradients of the model parameters ($\theta$) with respect to the loss function are calculated during SGD training, an attacker can compute gradients of the input pixels with respect to the classification output.
These gradients instruct the attacker on how to modify the image to change the model's classification.
In the following sections, we will briefly describe each of the adversarial attacks used in our experiments.

\paragraph{Fast Gradient Sign Method (FGSM)} \cite{Goodfellow2014ExplainingAH} uses the sign of the gradient of the loss function to determine the direction of the perturbation $\delta_x$. 
\[
    x' = x + \epsilon \times \text{sign}(\nabla \mathcal{L} (F(x|\theta),\widetilde{y}))
\]

Where $\epsilon$ is a small value which controls the magnitude of perturbation, chosen through trial and error.
To ensure a sufficiently small $\delta_x$, FGSM optimizes the $L_\infty$ distance between $x$ and $x'$, and thus tries to minimize the maximum change to any of the pixels.
It has been shown that with small values of $\epsilon$, most deep neural networks trained on CIFAR-10 and MNIST can easily be fooled \cite{Goodfellow2014ExplainingAH}.
FGSM is a particularly efficient attack method, as it requires only a single computation of the gradients of the input pixels.
However, FGSM does not always produce robust results.

\paragraph{Iterative Gradient Sign Method (IGSM)} \cite{Kurakin2016AdversarialEI} builds upon FGSM.
Starting with the original image, FGSM is applied iteratively until a satisfactory image is found, and, at each step, the generated adversarial image is clipped so that the generated image is within an $L_\infty$ $\epsilon$-neighborhood of the original image.
\[
x_0' = x, \qquad x_{N+1}' = \text{Clip}_{x,\epsilon} \Bigl\{x_N' + \alpha \times \text{sign}(\nabla \mathcal{L} (F(x|\theta),\widetilde{y}))  \Bigr\} 
\]
Here, $\alpha$ is the amount of perturbation added per iteration, which is usually very small compared to the $\epsilon$ used in FGSM.
This allows the attacker to find adversarial images which are closer to the original image than those found by FGSM~\cite{Kurakin2016AdversarialEI}.


\paragraph{Gradient Attack (GA)} is where a single step is taken, as in FGSM, but the step is taken in the exact direction of the gradient, rather than the direction of the element-wise sign of the gradient. The gradient step is normalized to have unit length.
\[
\mathcal{G} = \nabla \mathcal{L} (F(x|\theta),\widetilde{y}), \qquad x' = x + \epsilon \times \frac{\mathcal{G}}{\mathcal{||G||}_2} \qquad \text{where, $||.||_2$ denotes $L_2$ norm}
\]
\paragraph{Deep Fool}~\cite{MoosaviDezfooli2016DeepFoolAS} makes the simplifying assumption that the classifier to be targeted is a linear model which uses a hyperplane to separate the classes.
It tries to find an adversarial image by moving in the direction orthogonal to the linear decision boundary. 
Because the classifier is not actually linear, it is necessary to iterate this process until an adversarial image is found.
Deep Fool is an untargeted attack and optimizes for $L_2$ (PSNR) instead of $L_\infty$, unlike FGSM. 
Experiments reveal that perturbations generated by Deep Fool are less perceptible than those generated by other methods.
\paragraph{Jacobian-based Saliency Map Attack (JSMA)} \cite{Carlini2017TowardsET}
uses an adversarial saliency map, which indicates the pixels in the image that the adversary should increase in value in order to fool the classifier.
Let $\mathcal{J}(x)$, where  $\mathcal{J}_i(x) = \frac{\partial F_{i}(x)}{\partial x} $, be the Jacobian of the output of the classifier with respect to the image $x$ and $c'$ be the targeted class.
Then, the adversarial saliency map, $S(x_j,c')$, is given by: 
\[
 S(x_j,c') = \begin{cases} 
            0, & \text{if $\mathcal{J}_c'(x_j)<0$}$ or $\sum_{i \neq c'} \mathcal{J}_i(x_j) > 0 \\
           \mathcal{J}_{c'}(x_j) | \sum_{i \neq c'} \mathcal{J}_i(x_j) |, & \text{otherwise}
          \end{cases}
\]
The attacker tries to find the pixel {$x_j$} such that $S(x_j,c')$ is maximized. 
Then, that pixel is perturbed by $\epsilon$. 
This process is iterated until the model's prediction for $x$ is $c'$. 
JSMA produces alterations that are imperceptible to humans, but it is considerably slower than other techniques.

\paragraph{L-BFGS} \cite{Szegedy2013IntriguingPO} is one of the earliest proposed techniques to generate adversarial images.
The authors formulate the task of finding an adversarial image $x'$ given an image $x$ as a box-constrained optimization problem.
The constraint is the class label and the function being minimized is the $L_2$ distance between the image $x$ and the generated adversary $x'$.
In practice, it is hard to find a satisfactory minimum under such strict constraints. Instead, they reduce the minimization to:
\[
\text{minimize} \quad \alpha || x - x' ||_2^2 + \mathcal{L}((F(x'|\theta),c'),\quad \text{subject to } x' \in [0,1]^n 
\]
where $\alpha$ is an arbitrary value chosen by line search. 
This optimization is often carried out using L-BFGS, which is slower than other adversarial techniques.

\section{Defenses} \label{defenses}
Defenses against adversarial systems are either specific to a given attack or attack-agnostic. 
We briefly discuss some well-known defenses from both categories.

One of the techniques to defend against adversarial images is to include such images as part of the training set, thereby encouraging the model to classify the perturbed images with same label as the original image. 
This method is very inefficient as the space of possible perturbations is very large and it is not practical to train
a model long enough to become robust against all possible noise \cite{Tramr2017TheSO}. 
Another limitation of this method is that adversarial images trained to fool other networks are not defended against. 
It has been shown that adversarial images are transferable between models~\cite{Liu2016DelvingIT}, and such transferred images are generally difficult to defend against.
Adversarial systems which include random perturbations to the image before applying an attack technique are generally robust against
this kind of defense.

One of the most effective techniques towards defending against adversarial perturbations is called Defensive Distillation.
In this scheme, the outputs of the classifier are used to train another model, called the distilled model. 
The distilled model is trained to match the output of the last layer (softmax layer) of the original classifier.
This distilled model benefits from a smoother distribution of target outputs as compared to the all-or-nothing ground truth labels.
This model has been shown to be robust against some of the less efficient attacks.
However, this comes at the expense of training a new model.
Recently, quantization and other image processing techniques have been proposed as a first line of defense to counteract adversarial perturbations~\cite{liang2017detecting,aadityaprakash2018,guo2017countering}. 
These defenses significantly alter the image and render them unfit for general purpose use.
These techniques also reduce the classification accuracy of clean image, which is an undesirable effect of image transformations.

\begin{figure}
    \centering
     \includegraphics[width=0.46\textwidth,angle=0]{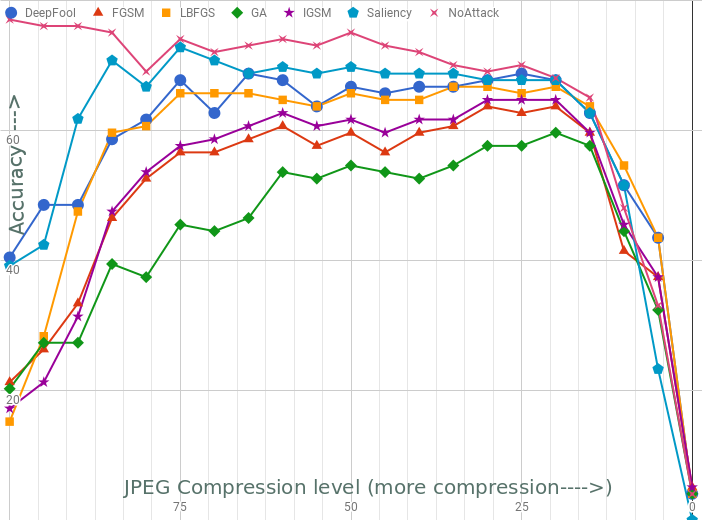}
     \includegraphics[width=0.52\textwidth,angle=0]{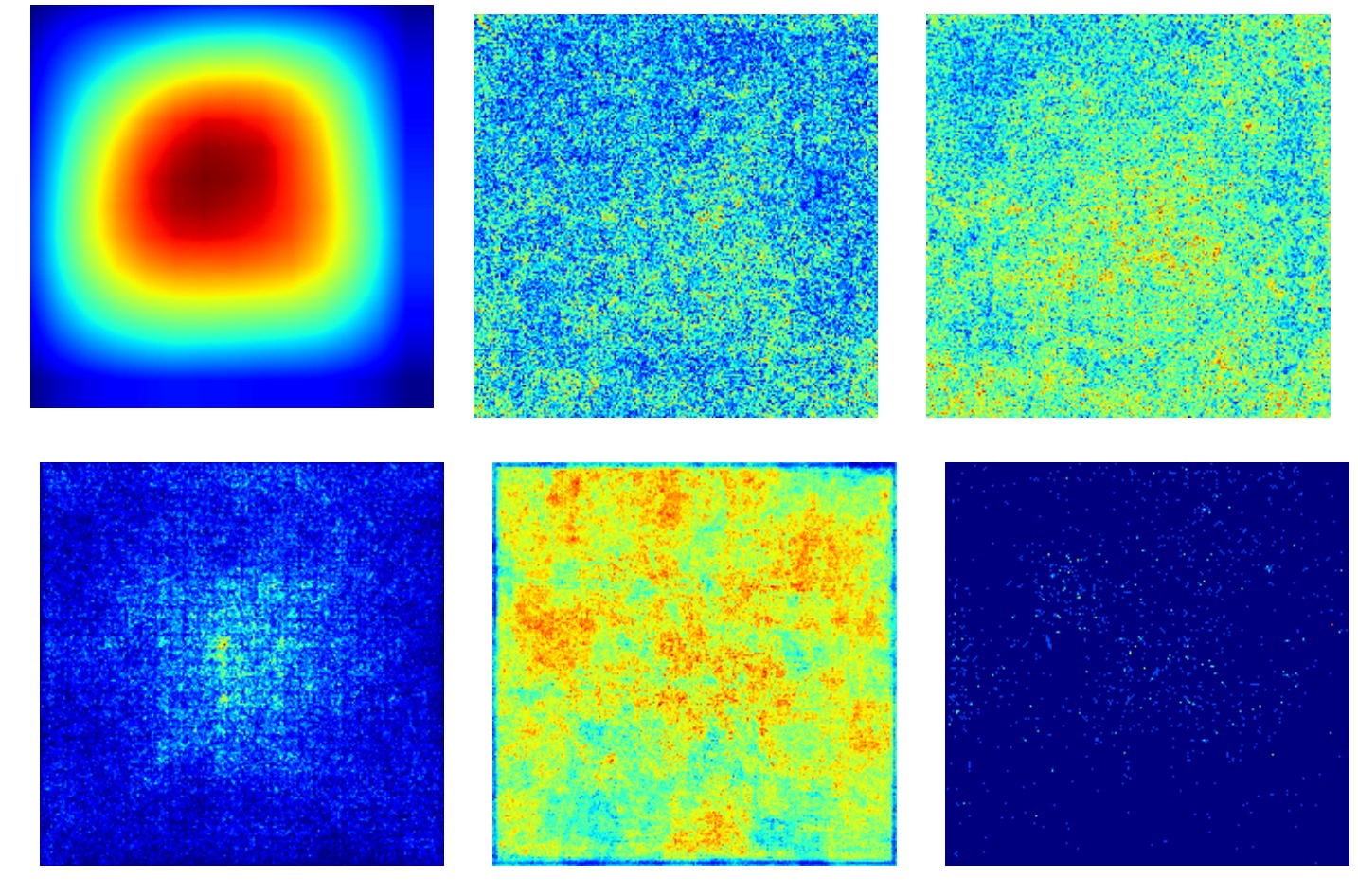}
     \label{fig:jpeg}
     \caption{(a) Classification accuracy on adversarial images is recovered as quantization increases, but drops when quantization becomes too aggressive. \textit{JPEG Compression level} refers to the scaling factor of the quantization matrix often denoted as Q. (b) Clockwise from top left; average MSROI saliency map for original images, average adversarial perturbations for five attack methods (IGSM, FGSM, JSMA, Deep Fool, GA).  Note that noise intensity is not concentrated in salient regions.}
 \end{figure}


\paragraph{Limitations of JPEG} - Recently, it has been shown that image transformations such as Gaussian blur, Gaussian noise, change of contrast and brightness, and image compression like JPEG are able to recover some portion of correct classifications when applied to adversarial images~\cite{Kurakin2016AdversarialEI}.
The impact of JPEG compression was further studied by~\cite{Dziugaite2016ASO} and~\cite{Das2017KeepingTB}.
Their studies were limited to less effective attack techniques, such as FGSM, but demonstrated the baseline efficacy of JPEG. 
We show that standard JPEG does not defend against newer and more robust attacks, and analyze some of the limitations of pure JPEG for adversarial defense.

JPEG's success in defending against some attack models is due to the quantization of high-frequency signals in the image.
The high frequencies will contain some portion of the perturbations added by the attacker, and quantization will lessen their impact.
However, adversarial perturbations in lower frequencies remain, and the more aggressive quantization which is required to remove these perturbations results in lower image quality, making it difficult to recover the original classification.



To maximize the effectiveness of our defense, while still preserving sufficient quality to allow for correct classification, we seek to strongly quantize those regions of the image which are not salient to the true class, while weakly quantizing salient regions.
We present a method for achieving this in the following section.

\section{Semantic Quantization}

Our approach employs a transformation $T(x)$ of a given image $x$ which quantizes the image in the frequency domain like standard JPEG.
However, unlike JPEG, we determine the salient regions of $x$ , i.e. regions for which the presence of an object is likely, and quantize those regions at higher bit-rates than non-salient regions.
We use a convolutional neural network to generate a saliency heat map, 
which guides quantization.

The center plot on Figure 1 shows the map of the perturbations added by an adversarial system (IGSM). The adversary has made modifications to every image region (except where the image is saturated), even though the elephant is in the center of the image. Our method takes advantage of this, and strongly quantizes the image blocks outside of the salient region. 

\begin{figure}[h]
\centering
     \includegraphics[width=0.3\textwidth,angle=0,scale=0.7]{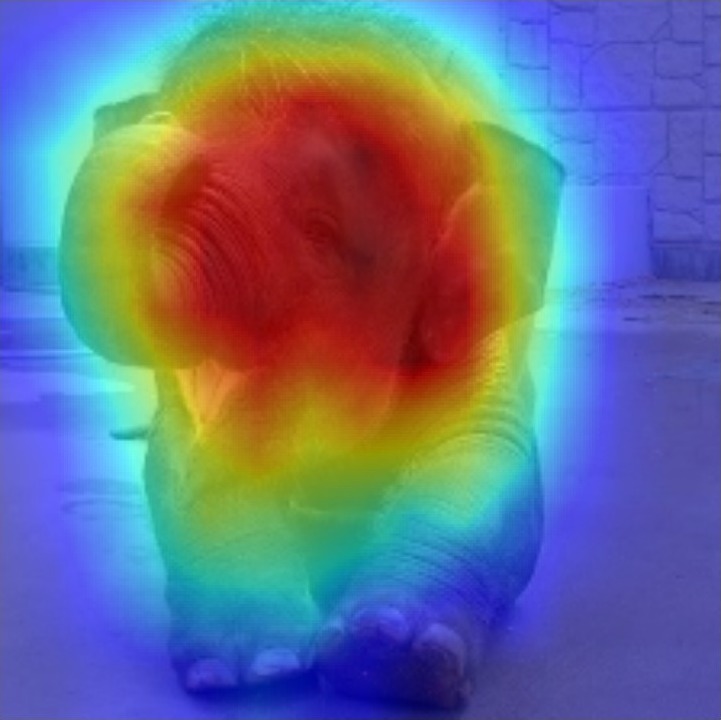}
     \includegraphics[width=0.3\textwidth,angle=0,scale=0.7]{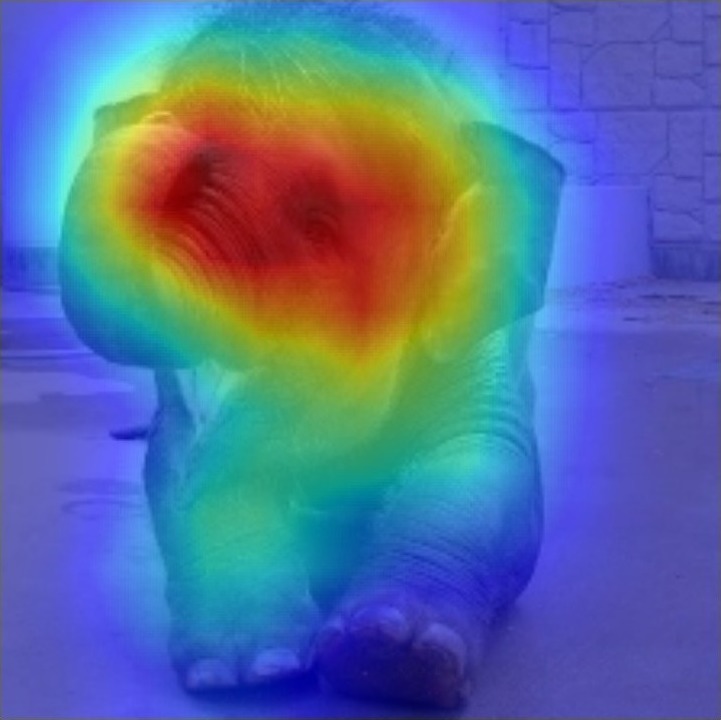}
     \includegraphics[width=0.3\textwidth,angle=0,scale=0.7]{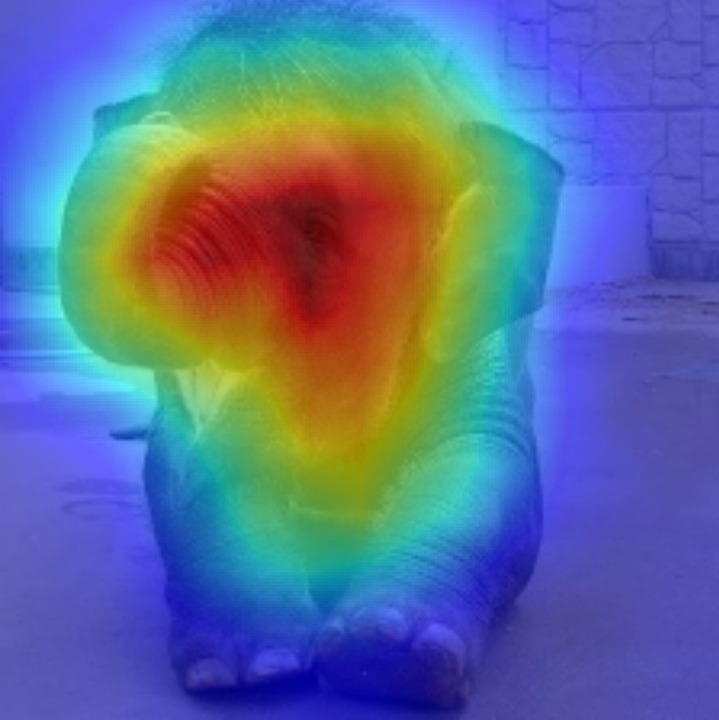}
     \label{fig:msroi}
          \caption{{\footnotesize : MSROI on original image, Center: MSROI, Right: Aug-MSROI, on adversarial image}}
 \end{figure}

Convolutional neural networks (CNNs) have been used to detect and locate objects in a given image \cite{Erhan2014ScalableOD,Zhao_2015_CVPR}. 
However, these works seek to generate a hard bounding box around the objects which does not take into consideration partial or occluded objects.
They are also limited in the number of classes they can detect, which ranges between $20$ to $40$.
Saliency detection techniques can solve these issues, but such techniques are limited in their ability to detect multiple objects, and the identified salient region may only contain a limited subset of the objects in the image.
To address these shortcomings, we utilize MSROI~\cite{Prakash2017SemanticPI}, a CNN designed to retrieve all salient regions and provide a soft boundary over the image.

MSROI generates a heat map based on its confidence over all classes and their corresponding activation values. 
At each layer, $l$, of the network, MSROI has independent filters for each class in the dataset. 
Thus, summing the filter responses of each class over their respective filters represents the confidence of that class. 
For an image $x$ at a given location of ($i,j$) and $C$ classes, MSROI map is defined as:
\[
M(i,j) = \sum_{c \in C} \begin{cases}
                        \sum_l f_l^c(i,j) & \text{if} \sum_{l} \sum_{x,y} f_d^c(i,j) > \mathcal{T} \\
                        0 & \text{otherwise}
                        \end{cases}
\]
\begin{figure}
\includegraphics[scale=0.4]{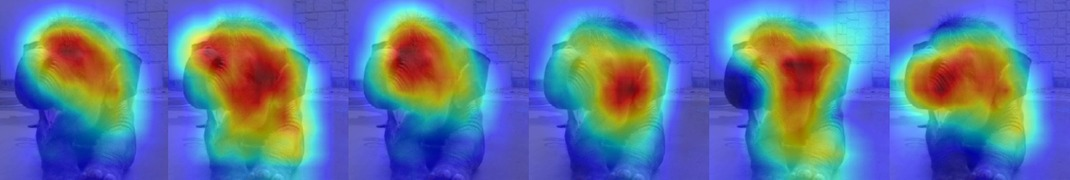}
     \label{fig:combination}
          \caption{Combination of various jittered inputs}
 \end{figure}

When the confidence of the model on a specific class is below a threshold $\mathcal{T}$ (determined through experimentation on validation data), the class is not represented. 

\section{Augmented MSROI}
MSROI was trained on natural images, but in this task it will be applied to an adversarial image. 
This means that the incorrect adversarial classification will be represented in the saliency map.
This presents an issue, as the adaptive quantization will try to preserve those areas which are salient to the incorrect classification. 
Thus, performing standard MSROI retains some adversarial noise which could have been quantized away.
To counteract this, we augment the MSROI algorithm by generating several small random perturbations of the image and the intermediate activations of the MSROI network, and calculating a saliency map $\widetilde{M}$ for each such perturbation:
\[
\widetilde{M}(i,j) = \sum_{c \in \hat{C}}  \sum_l (a_l^c(i ,j))  \text{\quad and \quad} a_{l+1}^c = f_l(a_l^c(i,j) +\Delta)
\]
where $\Delta$ is a random perturbation and $\hat{C}$ is subset of $C$ found similarly using $\mathcal{T}$ as shown with MSROI.
We take an average of $\widetilde{M}$ maps over several perturbations to generate a final map, which we term Augmented MSROI (Aug-MSROI).
This process of averaging over many perturbations gives softer localization which is less vulnerable to attack.
The number of perturbations to average over can be found empirically.  
Figure 4 shows several individual $\widetilde{M}$ maps which are combined to create the final heat map. 
In Figure 3 we show the difference between standard MSROI and Aug-MSROI. 
Most of the head of the baby elephant is retrieved in the Aug-MSROI heat map but not in the standard MSROI. 

Once the semantic heat map is obtained, we use the technique proposed in~\cite{Prakash2017SemanticPI}, which quantizes different $8\times8$ blocks at different levels and combines them to obtain a final JPEG image. 
This final image retains high quality in salient regions, and is heavily quantized in non-salient regions.
Strong quantization in non-salient regions is often sufficient to reverse the impact of adversarial perturbations, and high quality in salient regions allows recovery of the correct classification.

Our modifications of the standard MSROI algorithm result in only a modest increase in encoding complexity, and, like standard MSROI, the resulting compressed image can be decompressed by a standard off-the-shelf JPEG decoder.
Even though JSMA and L-BFGS attack selected pixels, the pixels changed by these attacks do not significantly correlate with the pixels selected by Aug-MSROI. Aug-MSROI extracts pixels belonging to the object while JSMA extracts pixels most likely to cause the change in image classification, which often lie in the negative space.


\section{Experiments}

We experimented with several adversarial attacks as described in Section~\ref{attacks}.
We used Residual Network (ResNet-50)~\cite{He2016DeepRL}, a state-of-the-art deep CNN as the classification model. 
Our test images are scaled and cropped to $256 \times 256$, and are drawn from the ImageNet dataset, which contains $1000$ classes.
All the attacks were parametrized to have RMSE distance of $0.02$ compared to the original image.
Aug-MSROI, like MSROI, requires a one-time off-line training of the model. Encoding of images can be done in parallel for several images on a GPU. Computation of Aug-MSROI map requires multiple perturbations for the same image and thus is somewhat more resource intensive than computing MSROI map. However, these passes can be computed in parallel as the final map is an average of individual maps. On a Titan X Maxwell GPU with 12 GB of memory encoding 300 images, each with five perturbations, took 5 seconds total; the decoding process remains unchanged from standard JPEG.

\paragraph{Metrics for comparison} Accuracy is the most common metric for assessing performance of deep neural networks on image classification tasks.
Most publications report `Top-5' accuracy, which is calculated by counting the number of instances for which the image's true label is in the top five predicted labels. 
For the purposes of adversarial attacks, we consider an attack to be successful if the model's top predicted class $c'$ for adversarial image $x'$ is not the same as the top predicted class for the original image $x$ i.e `Top-1' accuracy. 
This is a very generous success metric for the adversary, as the predicted class label for most of the adversarial images are within the `Top-5' of the original image even if not in `Top-1'.


\begin{figure}[h]
   \centering
     \includegraphics[width=0.24\textwidth,angle=0]{ori.pdf}
     \includegraphics[width=0.24\textwidth,angle=0]{adv.pdf}
     \includegraphics[width=0.24\textwidth,angle=0]{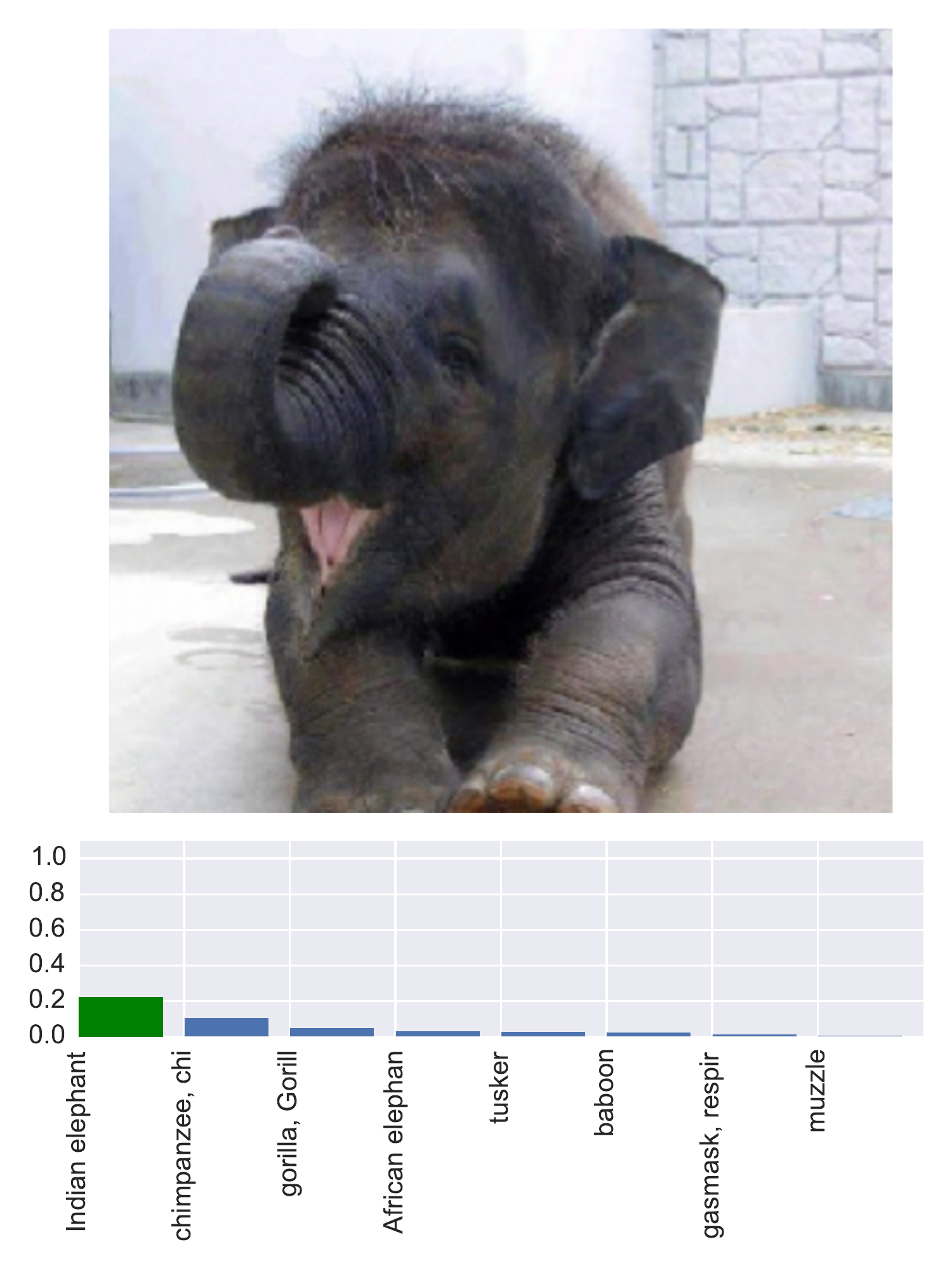}
     \includegraphics[width=0.24\textwidth,angle=0]{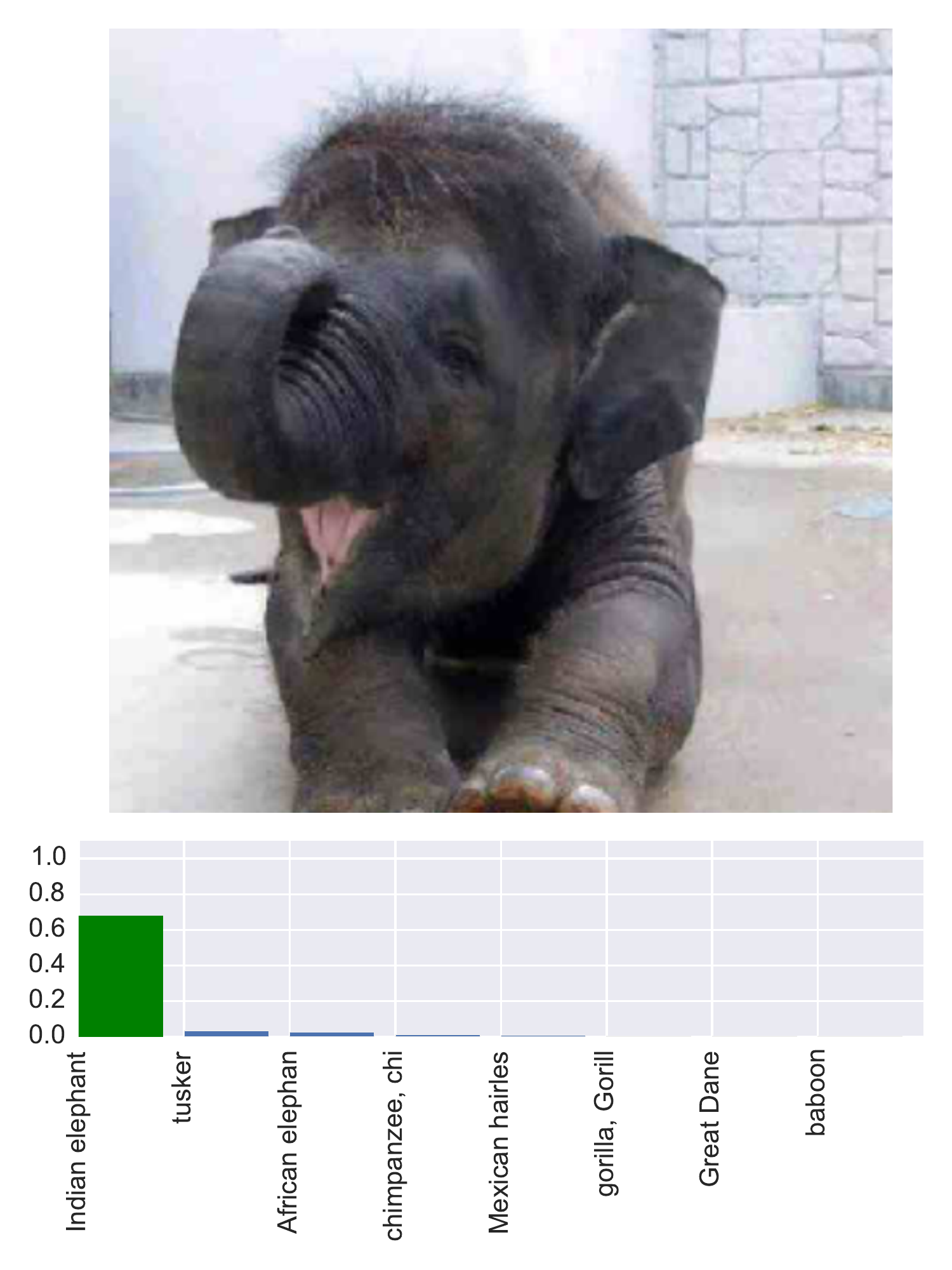}
              \caption{Original Image, Adversarial Image, JPEG, Aug-MSROI }
     \label{fig:compare}
 \end{figure}

 \begin{figure}[h]
     \centering
     \includegraphics[width=0.99\textwidth,angle=0]{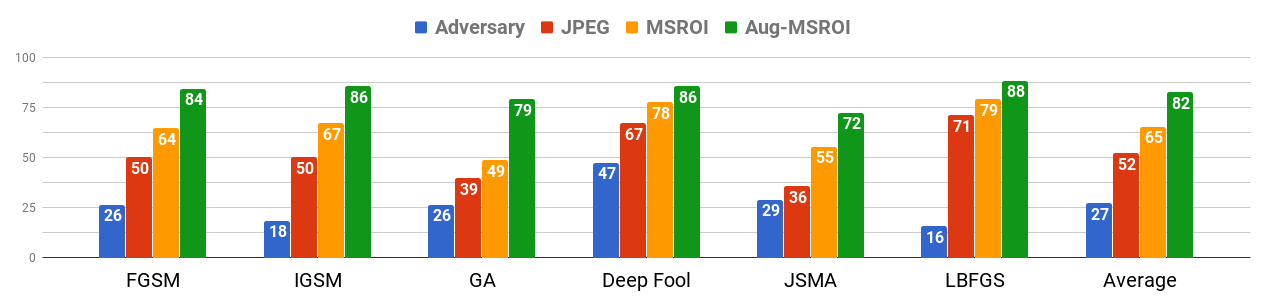}
     \caption{Classification accuracy of images on various attacks}
     \vspace{-5mm}
     \label{fig:numbers}
     \end{figure}

\section{Results}
Figure~\ref{fig:compare} shows a typical example of the results for qualitative comparison. In this case, while JPEG was also able to retrieve the original class, it does so with significantly lower confidence. 
Table 1 shows the quantitative results averaged across 300 images of ImageNet.

\begin{table}[H]
\centering
\label{tbl:results}
\begin{tabular}{rclll}
\multicolumn{1}{l}{} & \textbf{Accuracy$\uparrow$} & \textbf{MSE$\downarrow$} & \textbf{SSIM$\uparrow$} & \textbf{PSNR$\uparrow$} \\ \hline
\textbf{JPEG}        & 52                & 16.55        & 0.62          & 24.33         \\
\textbf{MSROI}       & 65                & 16.13        & 0.64          & 24.61         \\
\textbf{Aug-MSROI}   & 82                & 10.92        & 0.86          & 27.36        
\end{tabular}
\caption{Results averaged across $300$ images. }
\end{table}

The JPEG quantization level was chosen to achieve the highest accuracy. 
For MSROI, we applied the technique described in~\cite{Prakash2017SemanticPI} and set the quantization level to keep the final image quality as close as possible to the JPEG version. The results are shown in Figure 2(a). 
Aug-MSROI improved classification accuracy while maintaining the perceptual quality as evaluated by PSNR, MSE and SSIM~\cite{wang2004image}. 

\section{Conclusion}


We have presented Aug-MSROI, an augmentation of MSROI~\cite{Prakash2017SemanticPI}  to employ semantic JPEG compression as an effective defense against state-of-the-art adversarial attacks.
In our experiments, JPEG compression employing Aug-MSROI produces compressed images, which, when decompressed, have high visual quality while at the same time preserving the accuracy of classification by a neural network. 
A key advantage of our approach is that, with only a modest increase in encoding time, like standard MSROI, it produces compressed images that can be decompressed by any off-the-shelf JPEG decoder.

\section{Acknowledgement}
We would like to thank NVIDIA for donating the GPUs used for this research. 
We would also like to thank anonymous reviewers and Ryan Marcus (Brandeis University) for valuable feedback.

{\small
\bibliographystyle{ieee}
\bibliography{egbib}
}

\end{document}